\documentclass[pdflatex,sn-mathphys-num]{sn-jnl}

\usepackage{graphicx}%
\usepackage{multirow}%
\usepackage{amsmath,amssymb,amsfonts}%
\usepackage{amsthm}%
\usepackage{mathrsfs}%
\usepackage[title]{appendix}%
\usepackage{xcolor}%
\usepackage{textcomp}%
\usepackage{manyfoot}%
\usepackage{booktabs}%
\usepackage{algorithm}%
\usepackage{algorithmicx}%
\usepackage{algpseudocode}%
\usepackage{listings}%
\usepackage{comment}
\usepackage{physics}

\theoremstyle{thmstyleone}%
\theoremstyle{thmstyletwo}%

\theoremstyle{thmstylethree}%

\begin{document}

\title[Information-Theoretic Quality Metric of Low-Dimensional Embeddings]{Information-Theoretic Quality Metric of Low-Dimensional Embeddings}


\author*[1]{\fnm{Sebastián} \sur{Gutiérrez-Bernal}}\email{sebastian.gtz.brnl@tec.mx}

\author*[1]{\fnm{Hector} \sur{Medel Cobaxin}}\email{hmedel@tec.mx}

\author[1]{\fnm{Abiel} \sur{Galindo González}}\email{A00836342@tec.mx}

\affil*[1]{\orgname{Tecnologico de Monterrey}, \orgdiv{Escuela de Ingenieria y Ciencias}, \orgaddress{\street{Ave. Eugenio Garza Sada 2501 Sur, Col: Tecnologico}, \city{Monterrey}, \state{N.L.}, \country{Mexico}, \postcode{64700}}}

\abstract{In this work we study the quality of low-dimensional embeddings from an explicitly information-theoretic perspective. We begin by noting that classical evaluation metrics such as stress, rank-based neighborhood criteria, or Local Procrustes quantify distortions in distances or in local geometries, but do not directly assess how much information is preserved when projecting high-dimensional data onto a lower-dimensional space. To address this limitation, we introduce the Entropy Rank Preservation Measure (ERPM), a local metric based on the Shannon entropy of the singular-value spectrum of neighborhood matrices and on the stable rank, which quantifies changes in uncertainty between the original representation and its reduced projection, providing neighborhood-level indicators and a global summary statistic. To validate the results of the metric, we compare its outcomes with the Mean Relative Rank Error (MRRE), which is distance-based, and with Local Procrustes, which is based on geometric properties, using a financial time series and a manifold commonly studied in the literature. We observe that distance-based criteria exhibit very low correlation with geometric and spectral measures, while ERPM and Local Procrustes show strong average correlation but display significant discrepancies in local regimes, leading to the conclusion that ERPM complements existing metrics by identifying neighborhoods with severe information loss, thereby enabling a more comprehensive assessment of embeddings, particularly in information-sensitive applications such as the construction of early-warning indicators.}

\keywords{Manifold Learning, dimensionality reduction, embedding quality, entropy, singular values}

\maketitle

\section{Introduction}\label{Section: Introduction}
In high-dimensional spaces, data geometry deteriorates, implying that pairwise distances become less informative, neighborhoods degenerate, and many models see their performance collapse as the ambient dimension $D$ grows, a phenomenon often described as a manifestation of the curse of dimensionality \cite{bellman1961curse}. In consequence, both statistical estimation and learning algorithms, that rely on Euclidean distances or local neighborhood structure, may be affected. This motivates the search for representations in which the relevant structure of the data becomes more regular and accessible.\smallskip

With these considerations in mind, various studies and proposals have been developed\cite{gracia2014methodology}, mainly assuming that observations $x \in \mathbb{R}^D$ are concentrated near a low-dimensional manifold $\mathcal{M} \subset \mathbb{R}^D$ of intrinsic dimension $d \ll D$, and to construct a mapping $f : \mathbb{R}^D \to \mathbb{R}^d$ such that preserves salient structural properties of $\mathcal{M}$, including connectivity, curvature and intrinsic distances. This manifold-learning perspective underlies a broad family of nonlinear dimensionality-reduction methods, including geodesic-based approaches such as Isomap \cite{tenenbaum2000global}, locally linear methods such as LLE and its variants \cite{roweis2000nonlinear,zhang2004ltsa}, and spectral or kernel methods such as Laplacian Eigenmaps and Kernel PCA \cite{belkin2003laplacian,scholkopf1998kpca}. \smallskip

Therefore, it is extremely necessary to evaluate the quality of these low-dimensional representations. A variety of criteria have been proposed for this evaluation\cite{gracia2014methodology}, such as global measures like Kruskal's stress\cite{kruskal1964multidimensional}, Sammon's stress\cite{sammon2006nonlinear}, and Residual Variance\cite{tenenbaum2000global}, which quantify distortions of pairwise distances across the entire dataset. On the other hand, local criteria focus on neighborhood relations and assess changes in nearest neighbors, rank orders, and local shape through indices such as Trustworthiness, Continuity\cite{venna2006local}, Mean Relative Rank Error\cite{lee2009quality}, and Local Procrustes\cite{goldberg2009local}.\smallskip

However, these existing measures are defined purely in geometric terms and therefore quantify only how distances, ranks or local shapes change under the mappings, without having one that directly assesses how much information is lost or preserved when the projection is generated. This omission is particularly problematic when dimensionality reduction is used as a preprocessing step for sensitive tasks, such as reconstructing attractors from time series embeddings \cite{takens1981embedding,fraser1986mutual,cao1997practical} or building early-warning indicators for regime shifts in dynamical and financial systems \cite{huang2017imml}. In such settings, the informational content carried by the embedding is as critical as its geometric fidelity, in recent work on stable rank and geometry-preserving embeddings\cite{ eftekhari2018stabilizing} suggests that spectral properties of the data matrix can play a central role in capturing this content.\smallskip

This work is motivated precisely by that gap. Our goal is to complement existing geometric criteria with a quality measure that incorporates an explicitly informational point of view on the effect of dimensionality reduction. By linking local spectral structure to entropy, we aim to quantify not only how the manifold is geometrically distorted, but also how the uncertainty and effective number of spectral directions change when passing from the original representation to its low-dimensional embedding.


\section{Case of study: Information-based manifold learning for early warnings}\label{Section 2: Case of study}
The motivation of this study arises from Huang's \textit{et al.}\cite{huang2017imml} work, where a Manifold Learning methodology based on information criteria is proposed. This method takes a financial time series to reconstruct its phase space through Takens’ theorem\cite{takens1981embedding}, building the delay vectors
\begin{equation}\label{eq:delay-vectors}
X_t = (x_t, x_{t-\tau}, x_{t-2\tau}, \dots, x_{t-(m-1)\tau}),
\end{equation}

choosing the delay $\tau$ as the first minimum of the auto mutual information function \cite{fraser1986mutual} and the embedding dimension $m$ through Cao's method\cite{cao1997practical}, obtaining a matrix $X \in \mathbb{R}^{n \times m}$ that represents the reconstructed phase space in high dimension.\smallskip

Their idea centers in an information-theoretic perspective, in which instead of describing neighborhoods through Euclidean distances, each reconstructed state $X_i$ is modeled as a discrete probability distribution $p_i$, estimated through kernel density, where the similarity between states is measured with the symmetrized Kullback–Leibler divergence
\begin{align}
    h(P_i, P_j)
= \sum_{t=1}^m p_i^{(t)} \log\frac{p_i^{(t)}}{p_j^{(t)}}
+ \sum_{t=1}^m p_j^{(t)} \log\frac{p_j^{(t)}}{p_i^{(t)}},
\end{align}

whose values form the matrix $H$ that replaces the classical geometric neighborhoods. \smallskip

From this, they construct a Manifold Learning method analogous to LLE\cite{roweis2000nonlinear}, but with weights defined by the information metric, in which the low-dimensional embedding $Y$ is obtained by minimizing
\begin{align}
\min_{Y} \sum_{i=1}^n \left\lVert Y_i - \sum_{j=1}^n h_{ij} Y_j \right\rVert^2,
\end{align}

which reduces to an eigenvalue problem over $M = (I - H)^\top (I - H)$, whose $d$ eigenvectors associated with the smallest eigenvalues form the reduced coordinates.\smallskip

Finally, in the methodology they quantify the quality of the embedding using the Procrustes measure to compare the high-dimensional representation against the reduced one, using datasets from a stochastic simulation and the financial series of CSI800 and S\&P 500, where they show that their information-based method has much lower Procrustes error compared with other relevant methods such as Isomap, LLE, or KPCA.\smallskip

One of the things we highlight most about their work is its direct application, because once the manifold is reduced in dimensions, through clustering techniques and using Hidden Markov models, they detect regions associated with pre-crisis dynamics, which allows the reduced representation to be used as an early-warning tool. Furthermore, we consider their analysis of associating the curvature of the manifold with a global indicator of the underlying attractor's rigidity or resilience to be extremely relevant.\smallskip

However, we observe two relevant methodological limitations. First, the Procrustes metric measures global geometric preservation but is not sensitive to local distortions. Second, although the method is information-based, no metric was introduced to evaluate how well the embedding preserves divergence structure in the original space. Consequently, the comparative evaluation does not fully examine the method along the most relevant theoretical dimensions. This leads us to propose a local metric that can explicitly characterize aspects associated with the loss or preservation of information.


\section{Background and Related Work}


\subsection{Manifold Learning Methods}\label{Section: Manifold Learning Methods}
Manifold learning methods are commonly categorized according to the type of geometric structure they are designed to preserve. In this work, we adopt the view that, despite differences in implementation, these methods share the common objective of recovering a low-dimensional representation that preserves the intrinsic distances or neighborhood relationships of an underlying manifold embedded in a high-dimensional space. Accordingly, we group them into three dominant paradigms: local methods, global methods, and kernel or spectral methods, each reflecting a distinct interpretation of the manifold hypothesis and of which aspects of the geometry should be preserved. For comprehensive reviews of these approaches, we refer the reader to \cite{roweis2000nonlinear,tenenbaum2000global,zhang2004ltsa,belkin2003laplacian}.\smallskip

Local methods assume that we can approximate the manifold by a collection of tangent planes. Given data points $x_1,\dots,x_n$, we construct neighborhoods $N(i)$, often via $k$-nearest neighbors, and impose that the local geometric relations within each neighborhood are preserved after embedding. For example, in Locally Linear Embedding (LLE)\cite{roweis2000nonlinear} we compute reconstruction weights $w_{ij}$ such that 
\begin{align}
    x_i \approx \sum_{j \in N(i)} w_{ij} x_j, \quad \sum_{j \in N(i)} w_{ij} = 1,
\end{align}

and then seek a low-dimensional representation $Y$ satisfying
\begin{equation}
y_i \approx \sum_{j \in N(i)} w_{ij} y_j.
\end{equation}

By construction, the embedding $Y$ preserves local linear structure and remains consistent with the manifold's tangent geometry. Related approaches, such as Local Tangent Space Alignment (LTSA)\cite{zhang2004ltsa}, preserve higher-order local properties by aligning estimated tangent spaces rather than reconstruction weights. The Hessian Eigenmaps (HLLE) \cite{donoho2003hessian} goes one step further by estimating a second-order local representation that is sensitive to the curvature of the manifold in each neighborhood and use it to obtain a smoother low-dimensional embedding.\smallskip

Global methods focus on geodesic distances along the manifold. The most classical is Isomap \cite{tenenbaum2000global}, which approximates intrinsic distances by computing shortest paths on a neighborhood graph using Dijkstra’s algorithm:
\begin{equation}
d_{\mathcal{M}}(x_i, x_j) \approx \text{shortest path in graph } G.
\end{equation}

We then embed these estimated geodesic distances into a low-dimensional space via classical multidimensional scaling, solving an eigenvalue problem. Because geodesics capture long-range curvature of the manifold, Isomap preserves global structure but may be sensitive to noise or disconnected neighborhoods.\smallskip

Kernel or spectral methods interpret dimensionality reduction as constructing a nonlinear mapping by specifying a kernel function $k(x_i, x_j)$ that implicitly embeds the data into a high-dimensional feature space. Kernel PCA \cite{scholkopf1998kpca} finds principal components by solving
\begin{equation}
K \alpha = \lambda \alpha,
\end{equation}

where $K_{ij} = k(x_i, x_j)$. This allows linear methods to operate on nonlinear data by lifting them into a reproducing kernel Hilbert space. Closely related are graph Laplacian methods such as Laplacian Eigenmaps \cite{belkin2003laplacian}, which build a weighted adjacency matrix W and compute eigenvectors of the Laplacian
\begin{equation}
L = D - W.
\end{equation}

These methods preserve diffusion geometry and are grounded in connections between discrete Laplacians and the Laplace-Beltrami operator on smooth manifolds.\medskip

Together, these families of methods illustrate the spectrum of approaches to approximating the geometry of an unknown manifold; from enforcing consistency with local linear approximations, to reconstructing global geodesic structure, to expressing the manifold through kernels that encode nonlinear similarity. In all cases, dimensionality reduction is achieved by solving constrained optimization or spectral problems designed to preserve intrinsic manifold relationships rather than ambient Euclidean distances.


\subsection{Quality Metrics}\label{Section: Quality Metrics}
To quantify how well the quality of an \textit{embedding} is preserved after dimensionality reduction, it is natural to interpret such quality in terms of the geometric structure that is lost when we apply this mapping. In the work of Gracia \textit{et al.}\cite{gracia2014methodology}, a robust analysis of the state of the art of these metrics is presented, where it is shown, in a general manner, that most of them can be classified into local or global criteria.\smallskip

From the global perspective, there exist measures such as Kruskal's stress\cite{kruskal1964multidimensional}, Sammon stress\cite{sammon2006nonlinear}, or the residual variance\cite{tenenbaum2000global}, which indicate how well pairwise distances are preserved throughout the entire dataset. Meanwhile, metrics based on local criteria focus on neighborhood structure; some aim to evaluate the appearance and disappearance of neighbors, as well as changes in their relative order, such as Trustworthiness and Continuity\cite{venna2006local} or the Mean Relative Rank Error\cite{lee2009quality}; and others compare geometrically the shape of each neighborhood, such as Local Procrustes\cite{goldberg2009local}. \smallskip

A common feature of these criteria is that, despite their differences, they all assess the quality of the 
embedding through the geometric distortions introduced by dimensionality reduction, either through global distances or through changes in neighborhood structure. However, and considering the analysis in the case study of Section \ref{Section 2: Case of study}, we observe that these metrics do not allow us to directly evaluate how much information is retained throughout the process, which motivates the need to develop an additional criterion that explicitly incorporates this dimension of the analysis.\smallskip

In what follows, we focus on two representative local metrics that will serve as benchmarks in Section \ref{Section: Results}: the Mean Relative Rank Error (MRRE) and the Local Procrustes statistic.\smallskip

 MRRE\cite{lee2009quality} adopts a local distance-preservation approach, constructing a co-ranking matrix $\mathbf{Q} = q_{kl}$ that summarizes how distance ranks change between the original space and the low-dimensional embedding.\smallskip

From $\mathbf{Q}$ different types of neighborhood errors are identified according to the blocks defined in the Lee's work:
\begin{itemize}
    \item The block $\mathbb{L}\mathbb{L}_k$ counts the \emph{hard intrusions}, which are points that did not belong to the neighborhood in high dimension but enter it in the low-dimensional embedding.
    \item The block $\mathbb{U}\mathbb{R}_k$ counts the \emph{hard extrusions}, which are neighbors that existed in high dimension but cease to be neighbors in the low-dimensional embedding.
    \item And the block $\mathbb{U}\mathbb{L}_k$ groups the \emph{mild intrusions} and \emph{mild extrusions}, which are smaller rank changes in which, despite shifting, the points remain within the neighborhood.
\end{itemize}

This metric penalizes these errors through the following expressions:
\begin{align}
    W_n(K) = \frac{1}{H_K} \sum_{(k,l)\in \mathbb{U}\mathbb{L}_k \cup \mathbb{L}\mathbb{L}_k}   \frac{|k-l|}{l}q_{kl}, \quad \quad  W_v(K) = \frac{1}{H_K} \sum_{(k,l)\in \mathbb{U}\mathbb{L}_k \cup \mathbb{L}\mathbb{L}_k}   \frac{|k-l|}{l}q_{kl}, 
\end{align}

where 
\begin{align}
    H_K = N\sum_{k=1}^K \frac{|N -2k +1|}{k}
\end{align}

is a normalization factor.\smallskip

The Local Procrustes measure\cite{goldberg2009local}, in turn, indicates the degree of geometric preservation for each neighborhood, where centered neighbor matrices are used to remove translation effects:
\begin{align}
    \tilde{X}_i = (I-\frac{1}{k}\mathbf{1}\mathbf{1}^\top)X_i,\quad \tilde{Y}_i = (I-\frac{1}{k}\mathbf{1}\mathbf{1}^\top)Y_i,
\end{align}

and, with the SVD of $\tilde{X}_i^{\top}\tilde{Y}_i = U_iS_iV_i^{\top}$, a rotation matrix and a scaling factor are obtained:
\begin{align}
    A_i = U_i V_i^{\top}, \quad c_i=\frac{\text{tr}(S_i)}{\text{tr}(\tilde{Y}_i^{\top}\tilde{Y}_i)},
\end{align} 

with the purpose of aligning $\tilde{Y}_i$ and $\tilde{X}_i$ in the most optimal way. With this, the Procrustes statistic in its normalized conformal version is constructed:
\begin{align}
    R_{C,N}(X,Y)=\frac{1}{n}\sum_{i=1}^{n}\frac{G_{C_i}(X,Y)}{\bigl\lVert \tilde{X}_i\bigr\rVert_{F}^{2}}, \quad G_{C_i}(X,Y) = \bigl\lVert \tilde{X}_i - \tilde{Y}_i\,(c_i A_i^{\top}) \bigr\rVert_{F}^{2},
\end{align}

which averages the quality of the local adjustments across all neighborhoods.

\subsection{Information in terms of spectrum}\label{subsection: information}

Shannon \cite{shannon1948mathematical} introduces the notion of information as a measure of the uncertainty associated with the occurrence of an event prior to observing it, which we can more intuitively relate to the average amount of information we need to receive in order to know the state of the system. \smallskip

Mathematically, this average uncertainty is defined as the entropy of a discrete probability distribution
\begin{equation}\label{eq: Shannon entropy}
    H = -\sum_{j} p_j \log p_j,
\end{equation}

which measures the dispersion of the values $p_j$ and therefore \textit{the degree of lack of knowledge} about the system.\smallskip

In the context of comparing an object before and after subjecting it to a dimensionality reduction process, we see the need to measure not only how much distance was preserved but also how much of this uncertainty was modified or lost, because we cannot naturally guarantee that preserving distances also preserves information.\smallskip

As we know, in multivariate data, the eigenvalues of the covariance matrix measure variance along principal directions. These eigenvalues are proportional to the squared singular values of the centered data matrix, so the singular-value spectrum provides a compact summary of how variance is distributed across the dataset. We therefore define the entropy of a matrix $M$ by normalizing the squared singular values to obtain a probability distribution $p_j$ and evaluating the Shannon entropy in \eqref{eq: Shannon entropy}; quantifying how evenly the variance is spread across spectral directions: high values indicate that many directions contribute comparably, whereas low values indicate that only a few directions dominate. Consequently, when we apply a dimensionality-reduction method that eliminates degrees of freedom or concentrates variance into fewer directions, we observe a change in spectral entropy, which we can interpret as a variation of uncertainty with respect to the original representation. \smallskip

To illustrate this idea, let's consider a matrix $M\in \mathbb{R}^{d\times n}$ of algebraic rank $r_a = 3$, whose variance is distributed across three orthogonal singular directions. If we apply a dimensionality-reduction method that produces a new representation $\tilde{M}\in \mathbb{R}^{2\times n}$, then, by definition $r_a(\tilde{M})\leq 2$, implying that one of the three original spectral directions is lost or combined with the remaining ones, concentrating variance into fewer directions. In this sense, we can interpret the loss as a reduction in entropy, meaning a loss of information content of $\tilde{M}$ with respect to $M$.

\subsubsection{Stable Rank as a measure of spectrum}
The algebraic \textit{rank} of a matrix $M \in \mathbb{R}^{m\times n}$ is defined as the number of non-zero singular values, and it measures the dimension of the subspace spanned by its columns (i.e., the number of linearly independent directions contained in $M$). However, this rank depends in a non-continuous way on the entries of the matrix, implying that an arbitrarily small perturbation may change the number of non-zero singular values\cite{ipsen2025stable}, making it somewhat unstable in practice.\smallskip

In order to obtain an indicator that is continuous, differentiable, invariant under scaling, and robust to perturbations, the \textit{stable rank} is introduced\cite{rudelson2007sampling}
\begin{align}\label{eq: Stable Rank}
    r(M) = \frac{\lVert M \rVert_F^{2}}{\lVert M \rVert^{2}} 
= \frac{\sum_{j} \sigma_j(M)^{2}}{\sigma_1(M)^{2}},
\end{align}

where $\sigma_j(M)$ is the $j$-th singular value of $M$, and $1 \leq r(M) \leq \rank(M)$.\medskip

Additionally, we consider the proposal of Eftekhari \textit{et al.}\cite{eftekhari2018stabilizing}, in which a condition is established to determine when the \textit{delay-coordinate map} induces a \textit{stable embedding} of the attractor using \eqref{eq: Stable Rank}. From this perspective, stability is understood as the approximate preservation of distances between points on the attractor, leading to the conclusion that such stability requires the stable rank of the system to be proportional to the dimension of the attractor.\medskip

That said, if there exists a criterion that guarantees when a mapping is a stable embedding, then we consider that there must be an analogous notion of a \textit{stable projection} between two representations, one that, in the context of dimensionality reduction, is capable of quantifying the extent to which some information parameter is preserved after the reduction.

\section{Entropy Rank Preservation Measure (ERPM)}\label{Section 4. ERPM}

Starting from a matrix $M\in \mathbb{R}^{d\times n}$ with singular values $\sigma_1 \geq \sigma_2 \geq \cdots \geq \sigma_{r_a} >0$ and algebraic rank $r_a$, we associate to its spectrum a probability distribution
\begin{align}\label{eq: distribución espectral normalizada}
    p_j = \frac{\sigma_j^2}{\sum_{m=1}^{r_a} \sigma_m^2},
\end{align}

which describes the variation that exists along the spectral directions of $M$. We conveniently introduce a scaling parameter $\alpha_j = \sigma_j^2 / \sigma_1^2$, which satisfies
\begin{align}
    \sum_{j=1}^r \alpha_j = \frac{\sum_{j=1}^r \sigma_j^2}{\sigma_1^2} = r(M),
\end{align}

providing an alternative form of writing the stable rank defined in \eqref{eq: Stable Rank}. With this, it is evident that the expression \eqref{eq: distribución espectral normalizada} reduces to $p_j = \alpha_j/r(M)$, and that if we introduce these probabilities into \eqref{eq: Shannon entropy}, we obtain an entropy expression in terms of the stable rank:
\begin{align}\label{eq: Stable Rank Shanon Entropy}
    H(M) = \log{r(M)} - \varepsilon(M),
\end{align}

where 
\begin{align}
    \varepsilon(M) = \frac{1}{r(M)} \sum_{j=1}^r \alpha_j \log{\alpha_j}
\end{align}

is the term that reflects the internal distribution within the spectrum of matrix $M$.\smallskip

Now, let $X\in \mathbb{R}^{d\times n}$ be the high-dimensional embedding with $n$ observations, and $Y\in \mathbb{R}^{l\times n}$ its reduced representation. We define $\bar{X_i} = X_i - \frac{1}{k} X_i\mathbf{1}\mathbf{1}^{\top}$ and $X_i = \bigl[x_{i_j}\bigr]_{j=1}^k\in \mathbb{R}^{d\times k}$ as the centered matrix of the $k$ nearest neighbors for a point $x_i \in X$, constructing analogously the matrix $\bar{Y}_i$ in low dimension from the same set of neighbor indices.\smallskip

We substitute these matrices into \eqref{eq: Stable Rank Shanon Entropy}, obtaining uncertainty measures associated with the geometry of the neighborhood before and after dimensionality reduction. But, in order to quantify the change in the information of these representations, we compute the entropy difference, obtaining the \emph{Entropy Rank Preservation Measure}:
\begin{align}\label{eq: Entropy Rank Preservation Measure}
    \Delta H_i = H(\bar{Y}_i) - H(\bar{X}_i) = \log{ \qty(\frac{r(\bar{Y}_i)}{r(\bar{X}_i)})} + \varepsilon(\bar{X}_i) - \varepsilon(\bar{Y}_i).
\end{align}

It is important to highlight that, in general, any system will have an intrinsic indicator of uncertainty completely determined by its singular values. Therefore, we are interested in measuring the \emph{change} in this uncertainty, considering that even if the initial embedding has high or low entropy, what matters is that the variation $\Delta H_i$ is as close to zero as possible to consider that the projection generated by the reduction method is stable and shows practically no information loss.\smallskip

Finally, although the definition \eqref{eq: Entropy Rank Preservation Measure} is a local measure associated with each neighborhood, we can also define a global quantity that describes how much spectral information was modified throughout the embedding when the dimension was reduced. For this, we average the entropy change over all points:
\begin{align}\label{eq: global Entropy Rank Preservation Measure}
    R_{\Delta H} = \frac{1}{n} \sum_{i=1}^{n} \Delta H_i,
\end{align}

where positive values of this parameter indicate a global increase in data dispersion, negative values indicate a net loss of information by reducing the effective number of significant spectral directions, and values close to zero suggest a dimensionality reduction that preserves, on average, the spectral structure of the system. Complementarily, it is also necessary to analyze the distribution of the values $\Delta H_i$, as a favorable global average of $R_{\Delta H}$ may arise from a highly skewed or variable set of local behaviors. Such heterogeneity may produce neighborhoods in which the entropy changes sharply, despite the global measure suggesting that the overall distortion is relatively mild.

\section{Experimental Setup}

\subsection{Data, reduction methods, benchmarks and hyperparameters}
\begin{figure}[h]
\centering
\includegraphics[width=0.9\textwidth]{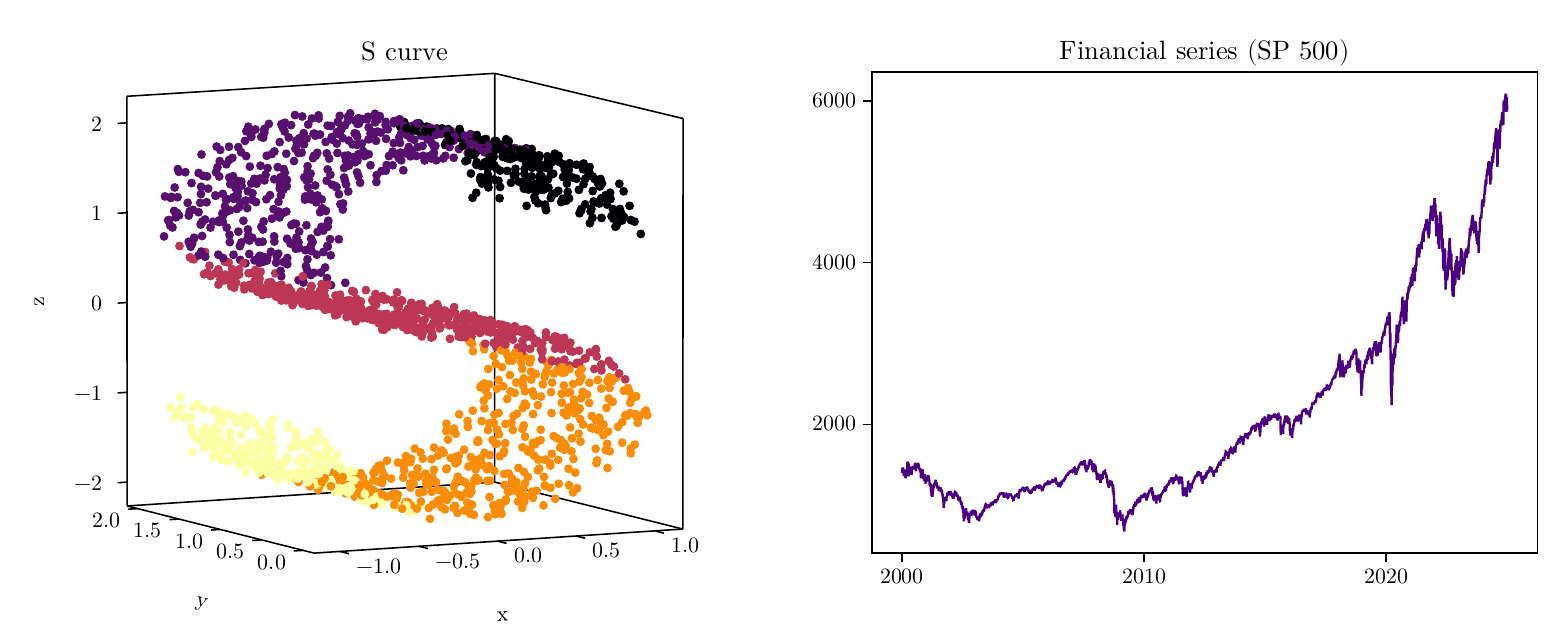}
\caption{Visualization of the data used in the metric evaluation.}\label{Fig: Datasets}
\end{figure}

We chose to employ two datasets, shown in Fig. \ref{Fig: Datasets}, in order to evaluate the behavior of the metrics under different scenarios. First, we used the S-curve with 2,000 points, a dataset widely studied in the Manifold Learning literature due to its nonlinear structure and its usefulness as a benchmark for dimensionality-reduction methods. Second, with the purpose of analyzing the information-theoretic aspects discussed in Section \ref{Section 2: Case of study}, we worked with the S\&P 500 financial series, collecting daily closing data from January 1, 2000 to January 1, 2025, from which we generated a Takens embedding following the same procedure as in the case study to estimate the optimal delay and dimension, obtaining values of 57 and 6, respectively.\smallskip

Regarding the dimensionality-reduction methods, we employed LLE, HLLE, Isomap, and KPCA with quadratic kernel, as described in Section \ref{Section: Manifold Learning Methods}. We also used PCA as a linear baseline, which orders orthogonal directions according to their variance, discards those associated with the smallest variance, and linearly projects the data into a lower-dimensional space. To benchmark the ERPM introduced in Section \ref{Section 4. ERPM}, we used the MRRE and Local Procrustes measures described in Section \ref{Section: Quality Metrics}, which allowed us to compare the resulting embeddings from complementary perspectives.\smallskip

With respect to the hyperparameters associated with the local character of the metrics, we analyzed their behavior as a function of the number of neighbors $k$ by performing a sweep over the interval $k \in [1,20]$. For the dimensionality-reduction methods that explicitly depend on this parameter, we used the value $k=k_{\max}$, thus ensuring a consistent comparison across techniques. Finally, in the last analysis of the following section, we fixed $k=15$ based on the onset of the stabilization period, whose analysis we discuss in greater detail later.

\section{Results}\label{Section: Results}

\subsection{Global Correlation with Benchmarks}
\begin{figure}[h]
\centering
\includegraphics[width=0.9\textwidth]{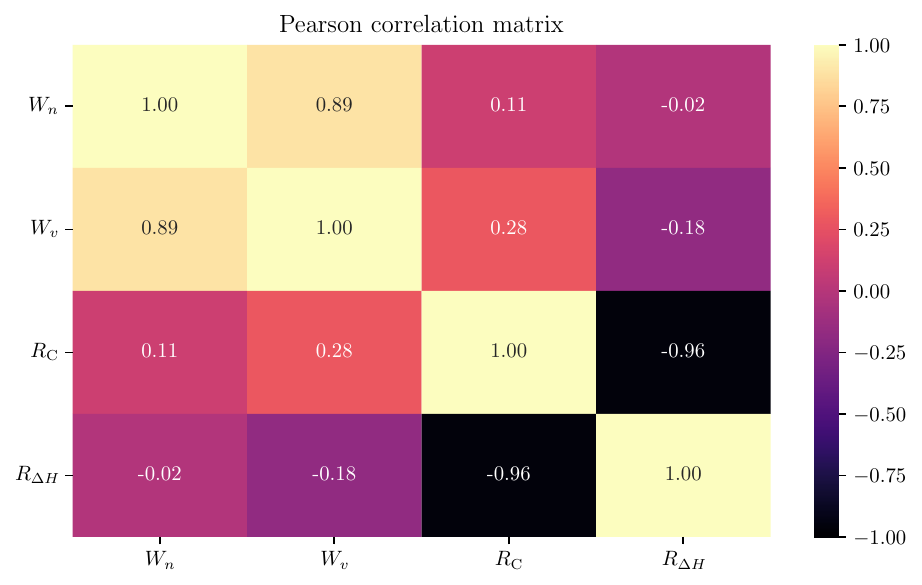}
\caption{Correlation plot between quality metrics.}\label{Fig: Correlation plot}
\end{figure}

We begin by analyzing the relationship between the ERPM and the benchmark metrics based on the correlation plot in Fig. \ref{Fig: Correlation plot}, where we observe that there is practically no correlation between distance-based metrics and metrics based on geometric and spectral properties, since both $W_n$ and $W_v$ exhibit extremely low correlation coefficients with $R_C$ and $R_{\Delta H}$, which suggests that preserving distances does not directly imply preserving the geometric structure nor the information content of the embedding. Therefore, if we want a more robust analysis of projection quality, we must treat distance metrics and spectral metrics as complementary indicators.\smallskip

On the other hand, we observe that the correlation between $R_C$ and $R_{\Delta H}$ is extremely high in absolute terms (the negative sign is due to their respective value ranges), which, from a relational perspective implies that both indicators consistently display the same type of local degradation in the embedded representation when viewed through their global measures. \smallskip

\begin{figure}[h]
\centering
\includegraphics[width=0.9\textwidth]{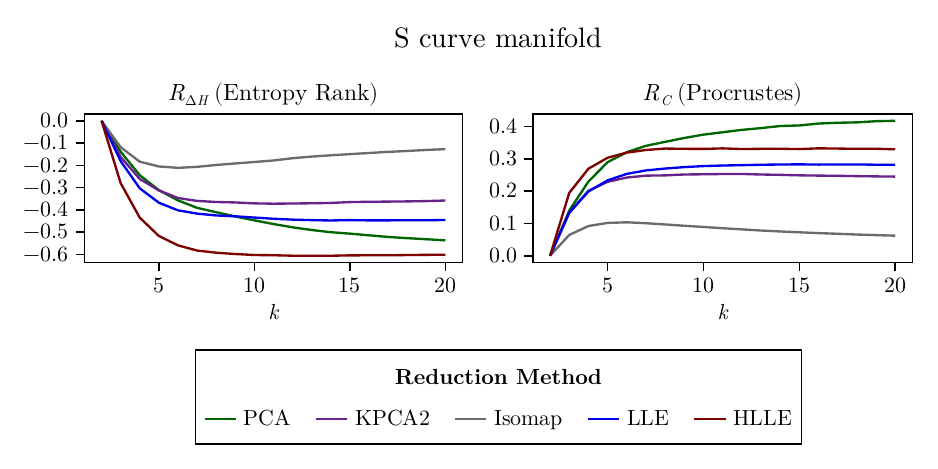}
\caption{Metrics comparison for the manifold dataset.}\label{Fig: Manifold metrics comparison}
\end{figure}

Considering this, and in order to further examine the correlation between both metrics, in Fig. \ref{Fig: Manifold metrics comparison} we analyze in a more granular manner the behavior of $R_{\Delta H}$ and $R_C$ for the S-curve dataset, where we first observe that both curves display a relatively sharp drop for small values of $k$, indicating a certain degree of local instability in the initial neighborhoods. Although the behavior becomes more stable as $k$ increases, global performance continues deteriorating for all reduction methods except Isomap. Furthermore, the ERPM tends to take more extreme values than Local Procrustes, showing that this metric penalizes more severely. However, at the level of overall trends, the profiles of $R_{\Delta H}$ and $R_C$ remain clearly aligned for all methods, which reinforces the strong relationship already identified in the correlation plot.\smallskip

\begin{figure}[h]
\centering
\includegraphics[width=0.9\textwidth]{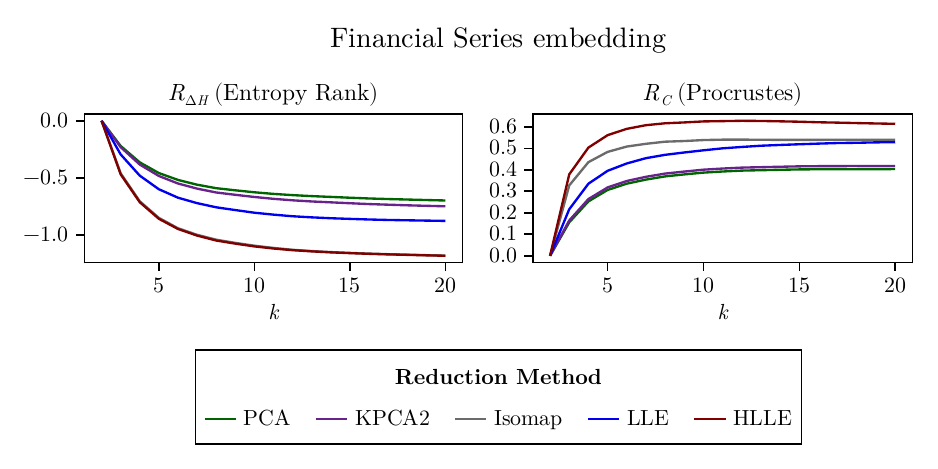}
\caption{Metrics comparison for the financial dataset.}\label{Fig: Financial series metrics comparison}
\end{figure}

If we repeat the same analysis for the financial data, shown in Fig. \ref{Fig: Financial series metrics comparison}, we find a behavior similar to that observed in the manifold, since both metrics exhibit a stabilization period followed by progressive deterioration as $k$ increases. In this case, however, PCA and KPCA2 are the methods that present the least negative values of $R_{\Delta H}$ and the smallest values of $R_C$, while LLE, Isomap, and HLLE display more severe losses or distortions. It is also relevant that, compared to the S-curve, the financial embedding produces notably more negative values in the entropy metric and substantially higher values in the Procrustes metric, an aspect consistent with the intrinsically noisy geometry of the financial embedding and which supports the usefulness of the ERPM as a specific indicator to characterize regimes of spectral information loss.\smallskip

\subsection{Local Analysis from a Neighborhood Perspective}

\begin{figure}[h]
\centering
\includegraphics[width=0.9\textwidth]{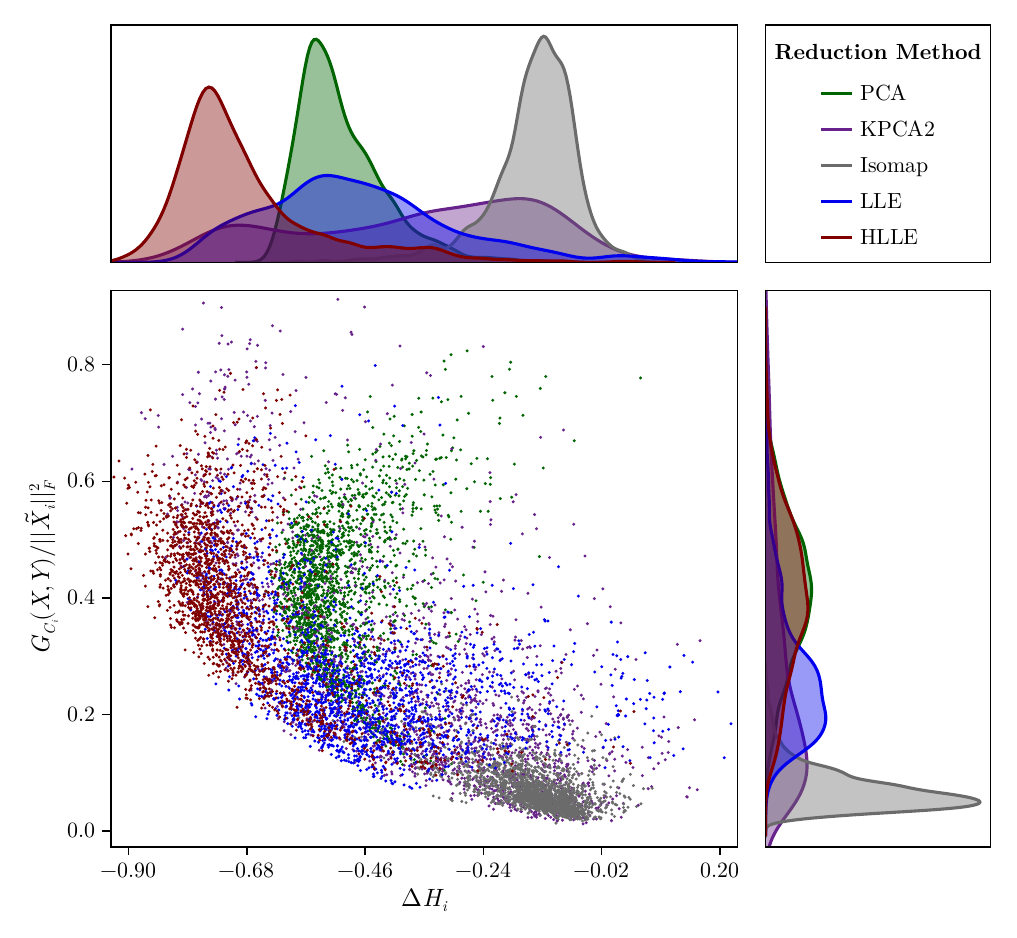}
\caption{Manifold data jointplot of the results of Local Procrustes (y-axis of the scatter plot located in the center and density plot on the right) and of ERPM (x-axis of the scatter plot located in the center and density plot at the top).}\label{Fig: Manifold Join Plot Procrustes-ERPM}
\end{figure}

Now, recalling what we mentioned in Section \ref{Section 4. ERPM} about the importance of analyzing not only the average of the metric, but also the full distribution of its values, we proceed to study the joint behavior of the ERPM and Local Procrustes, beginning with the distributions corresponding to the S-curve dataset shown in Fig. \ref{Fig: Manifold Join Plot Procrustes-ERPM}, where we observe that, at a global level, the central cloud of points follows a clearly decreasing trend: regions with greater information loss tend to be associated with higher Procrustes errors, while neighborhoods with ERPM values close to zero tend to display small Procrustes errors. This supports the strong correlation discussed throughout this Section, since methods that better preserve the global geometry also tend to preserve spectral entropy more effectively. However, the relationship ceases to be almost one-to-one when we analyze specific regions of the diagram, since for intermediate Procrustes values between $0.3$ and $0.5$, the range of $\Delta H_i$ extends from moderate values close to $-0.2$ up to very severe losses on the order of $-0.9$.\smallskip

If we analyze the marginal densities of the jointplot, we see that the peaks of the $\Delta H_i$ distributions are well separated across methods, with HLLE clearly located further to the left, indicating much more intense entropy loss and more concentrated data, though accompanied by several outliers that increase dispersion; but as we move to PCA, KPCA2, and LLE, the distributions progressively shift to the right, approaching less negative values of $\Delta H_i$, yet retaining distinguishable density profiles. Finally, Isomap presents a sharply concentrated distribution very close to zero, indicating that, in general terms, this method barely loses information. In contrast, the densities associated with Procrustes error are much more overlapped, considering that HLLE, PCA, and KPCA2 display very similar distributions, while Isomap stands out by accumulating most of its values near zero. With this we see that, from a purely geometric perspective, several methods may appear almost indistinguishable, but with entropy-change profiles that differ considerably. \smallskip

\begin{figure}[h]
\centering
\includegraphics[width=0.9\textwidth]{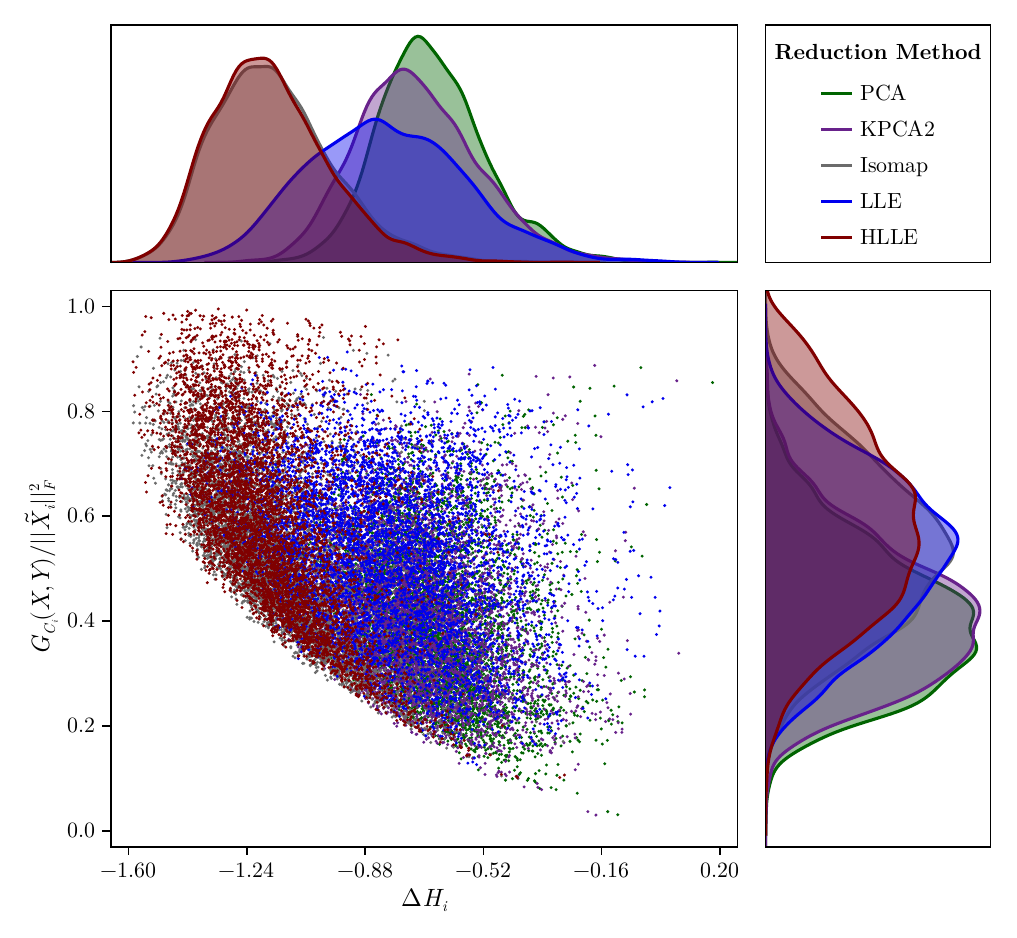}
\caption{Financial data jointplot of the results of Local Procrustes (y-axis of the scatter plot located in the center and density plot on the right) and of ERPM (x-axis of the scatter plot located in the center and density plot at the top).}\label{Fig: Finance Join Plot Procrustes-ERPM}
\end{figure}

When analyzing now the distributions corresponding to the financial data, shown in Fig. \ref{Fig: Finance Join Plot Procrustes-ERPM}, we recall that this dataset does not organize into a clean, low-dimensional manifold, so all methods face greater difficulties in producing high-quality embeddings. In this context, the scatter cloud again displays a sharply decreasing global trend, which continues to justify the correlation between the averages of $R_{\Delta H}$ and $R_C$. However, the cloud is now denser and the contributions of the different methods overlap more substantially, indicating that dimensionality-reduction schemes behave more similarly for this type of data. \smallskip

In the ERPM densities we observe that the methods share a broader distribution clearly skewed to the left, with peaks located closer together than in the S-curve case, although significant differences still exist. HLLE continues to occupy the most negative region, now accompanied by Isomap, whose distributions become comparable in terms of entropy loss. PCA and KPCA2, for their part, display less negative values and occupy intermediate positions, but with elongated left tails indicating neighborhoods with very severe information losses. Regarding the Procrustes error densities, overlap is even more pronounced than in the manifold dataset, since all distributions concentrate within a similar band and differ only by small shifts, such that methods showing substantial differences in the ERPM may appear almost equivalent when evaluated exclusively with Procrustes. Taken together, the shape of the distributions reveals heterogeneous regimes in which the local representation is strongly degraded, even though mean-based metrics may suggest a relatively homogeneous behavior. \smallskip

Based on these results, we interpret that, although we may expect a high correlation between the averages of both metrics, a description based solely on averages can be strongly biased, considering that at the neighborhood level the relationship between $R_{\Delta H}$ and $R_C$ weakens, since we may find representations in which we obtain similar geometric fits in terms of Procrustes but with very different information contents. Therefore, the central contribution of the ERPM does not lie only in its global average value, but in the complete structure of its distribution; and if we return to the case study in Section \ref{Section 2: Case of study} and the analysis carried out on the financial series, the pathological neighborhoods with very high entropy losses, despite presenting a moderate global average, are precisely the local structures we aim to identify and, insofar as possible, avoid to obtain higher-quality embeddings.

\section{Discussion}

In this work, by analyzing the quality of embeddings from geometric and informational perspectives, we find that, although distance preservation is an indispensable requirement in dimensionality-reduction procedures, it is insufficient to fully characterize the quality of the projection, since our results indicate that metrics based exclusively on distances can be almost independent from those that capture the geometric structure or the spectral information content of the embedding, and therefore these criteria must be treated as complementary rather than interchangeable indicators. \smallskip

This point is particularly relevant given the growing number of applications of dimensionality-reduction methods in sensitive contexts, such as the construction of Early Warning indicators for regime shifts or financial crashes discussed in Section \ref{Section 2: Case of study}. In these scenarios, the quality of the information preserved by the embedding is ultimately what supports the most consequential decisions. Motivated by this, we introduced the \emph{Entropy Rank Preservation Measure} (ERPM), a novel metric based on Shannon entropy associated with the local spectrum of the neighborhood matrices, which allows us to quantify, in terms of loss or gain of uncertainty, how much information content is preserved when we move from the original representation to its low-dimensional projection. \smallskip

The analysis also revealed the need to evaluate embedding quality from an explicitly local perspective, since although many metrics in the literature are defined at the neighborhood level, their usual use is restricted to global measures, which can hide pathological behaviors and lead to biased conclusions. By studying the full distribution of the local values of the ERPM, we observed that seemingly favorable averages can coexist with subsets of neighborhoods exhibiting extremely severe information losses, particularly in financial data with intrinsically noisy geometry. \smallskip

Taken together, these findings lead us to conclude that the central contribution of the ERPM does not lie solely in its global mean value, but in the framework it offers for analyzing in detail the local structure of information loss. By combining this metric with geometric criteria such as Local Procrustes and with distance-preservation measures such as MRRE, we obtain a much more holistic scheme for evaluating the quality of embeddings, capable of identifying particularly unstable regions and guiding the selection of dimensionality-reduction methods in applications where parameter preservation is critical.

\bibliography{sn-bibliography}

@article{shannon1948mathematical,
  title={A mathematical theory of communication},
  author={Shannon, Claude E},
  journal={The Bell system technical journal},
  volume={27},
  number={3},
  pages={379--423},
  year={1948},
  publisher={Nokia Bell Labs}
}

@book{bellman1961curse,
  title     = {Adaptive Control Processes: A Guided Tour},
  author    = {Bellman, Richard},
  year      = {1961},
  publisher = {Princeton University Press}
}

@article{eftekhari2018stabilizing,
  title={Stabilizing embedology: Geometry-preserving delay-coordinate maps},
  author={Eftekhari, Armin and Yap, Han Lun and Wakin, Michael B and Rozell, Christopher J},
  journal={Physical Review E},
  volume={97},
  number={2},
  pages={022222},
  year={2018},
  publisher={APS}
}

@article{zhang2004ltsa,
  title={Principal manifolds and nonlinear dimensionality reduction via local tangent space alignment},
  author={Zhang, Zhenyue and Zha, Hongyuan},
  journal={SIAM Journal on Scientific Computing},
  volume={26},
  number={1},
  pages={313--338},
  year={2004},
  publisher={SIAM}
}

@inproceedings{belkin2003laplacian,
  title={Laplacian eigenmaps for dimensionality reduction and data representation},
  author={Belkin, Mikhail and Niyogi, Partha},
  booktitle={Advances in Neural Information Processing Systems},
  volume={15},
  pages={585--591},
  year={2003}
}

@article{scholkopf1998kpca,
  title={Nonlinear component analysis as a kernel eigenvalue problem},
  author={Sch{\"o}lkopf, Bernhard and Smola, Alexander and M{\"u}ller, Klaus-Robert},
  journal={Neural Computation},
  volume={10},
  number={5},
  pages={1299--1319},
  year={1998},
  publisher={MIT Press}
}

@incollection{takens1981embedding,
  title={Detecting strange attractors in turbulence},
  author={Takens, Floris},
  booktitle={Dynamical Systems and Turbulence},
  pages={366--381},
  year={1981},
  publisher={Springer}
}

@article{fraser1986mutual,
  title={Independent coordinates for strange attractors from mutual information},
  author={Fraser, Andrew M and Swinney, Harry L},
  journal={Physical Review A},
  volume={33},
  number={2},
  pages={1134--1140},
  year={1986}
}

@article{ipsen2025stable,
  title={Stable rank and intrinsic dimension of real and complex matrices},
  author={Ipsen, Ilse CF and Saibaba, Arvind K},
  journal={SIAM Journal on Matrix Analysis and Applications},
  volume={46},
  number={3},
  pages={1988--2007},
  year={2025},
  publisher={SIAM}
}

@article{rudelson2007sampling,
  title={Sampling from large matrices: An approach through geometric functional analysis},
  author={Rudelson, Mark and Vershynin, Roman},
  journal={Journal of the ACM (JACM)},
  volume={54},
  number={4},
  pages={21--es},
  year={2007},
  publisher={ACM New York, NY, USA}
}

@article{gracia2014methodology,
  title={A methodology to compare dimensionality reduction algorithms in terms of loss of quality},
  author={Gracia, Antonio and Gonz{\'a}lez, Santiago and Robles, Victor and Menasalvas, Ernestina},
  journal={Information Sciences},
  volume={270},
  pages={1--27},
  year={2014},
  publisher={Elsevier}
}

@article{kruskal1964multidimensional,
  title={Multidimensional scaling by optimizing goodness of fit to a nonmetric hypothesis},
  author={Kruskal, Joseph B},
  journal={Psychometrika},
  volume={29},
  number={1},
  pages={1--27},
  year={1964},
  publisher={Springer-Verlag}
}

@article{sammon2006nonlinear,
  title={A nonlinear mapping for data structure analysis},
  author={Sammon, John W},
  journal={IEEE Transactions on computers},
  volume={100},
  number={5},
  pages={401--409},
  year={2006},
  publisher={Ieee}
}

@article{tenenbaum2000global,
  title={A global geometric framework for nonlinear dimensionality reduction},
  author={Tenenbaum, Joshua B and Silva, Vin de and Langford, John C},
  journal={science},
  volume={290},
  number={5500},
  pages={2319--2323},
  year={2000},
  publisher={American Association for the Advancement of Science}
}

@article{lee2009quality,
  title={Quality assessment of dimensionality reduction: Rank-based criteria},
  author={Lee, John A and Verleysen, Michel},
  journal={Neurocomputing},
  volume={72},
  number={7-9},
  pages={1431--1443},
  year={2009},
  publisher={Elsevier}
}

@article{venna2006local,
  title={Local multidimensional scaling},
  author={Venna, Jarkko and Kaski, Samuel},
  journal={Neural Networks},
  volume={19},
  number={6-7},
  pages={889--899},
  year={2006},
  publisher={Elsevier}
}

@article{goldberg2009local,
  title={Local procrustes for manifold embedding: a measure of embedding quality and embedding algorithms},
  author={Goldberg, Yair and Ritov, Ya’acov},
  journal={Machine learning},
  volume={77},
  number={1},
  pages={1--25},
  year={2009},
  publisher={Springer}
}

@article{cao1997practical,
  title={Practical method for determining the minimum embedding dimension of a scalar time series},
  author={Cao, Liangyue},
  journal={Physica D: Nonlinear Phenomena},
  volume={110},
  number={1-2},
  pages={43--50},
  year={1997},
  publisher={Elsevier}
}

@article{roweis2000nonlinear,
  title={Nonlinear dimensionality reduction by locally linear embedding},
  author={Roweis, Sam T and Saul, Lawrence K},
  journal={science},
  volume={290},
  number={5500},
  pages={2323--2326},
  year={2000},
  publisher={American Association for the Advancement of Science}
}

@article{huang2017imml,
  title={Nonlinear manifold learning for early warnings in financial markets},
  author={Huang, Yan and Kou, Gang and Peng, Yi},
  journal={European Journal of Operational Research},
  volume={258},
  number={2},
  pages={692--702},
  year={2017},
  publisher={Elsevier}
}

@article{donoho2003hessian,
  title={Hessian eigenmaps: Locally linear embedding techniques for high-dimensional data},
  author={Donoho, David L and Grimes, Carrie},
  journal={Proceedings of the National Academy of Sciences},
  volume={100},
  number={10},
  pages={5591--5596},
  year={2003},
  publisher={The National Academy of Sciences}
}

\end{document}